\documentclass[conference]{IEEEtran}

\usepackage[numbers,sort&compress]{natbib}

\usepackage{color,array,amsthm}
\usepackage{amssymb}        

\usepackage{graphicx}
\usepackage{subfigure}
\usepackage{amsmath}
\usepackage{multirow}
\usepackage{multicol}
\usepackage{svg}
\usepackage{soul} 
\usepackage{amsmath}
\usepackage{tcolorbox}
\usepackage{xcolor}
\usepackage{color}
\usepackage{pifont}
\usepackage{booktabs}
\usepackage{threeparttable} 
\ifCLASSINFOpdf
\else
\fi
\begin{document}
%
\title{CalibFusion: Transformer-Based Differentiable Calibration for Radar-Camera Fusion Detection in Water-Surface Environments}

\author{Yuting Wan, Liguo Sun, Jiuwu Hao, Pin LV~\IEEEmembership{Member,~IEEE,}
\thanks{Yuting Wan and Jiuwu Hao are with the School of Artificial Intelligence, University of Chinese Academy of Sciences, Beijing, China.}%
\thanks{Liguo Sun and Pin Lv are with the Institute of Automation, Chinese Academy of Sciences, Beijing, China.}%

}


%


\maketitle

\begin{abstract}
Millimeter-wave (mmWave) Radar--Camera fusion improves perception under adverse illumination and weather, but its performance is sensitive to Radar--Camera extrinsic calibration: residual misalignment biases Radar-to-image projection and degrades cross-modal aggregation for downstream 2D detection. Existing calibration and auto-calibration methods are mainly developed for road and urban scenes with abundant structures and object constraints, whereas water-surface environments feature large textureless regions, sparse and intermittent targets, and wave-/specular-induced Radar clutter, which weakens explicit object-centric matching.

We propose \textit{CalibFusion}, a calibration-conditioned Radar--Camera fusion detector that learns implicit extrinsic refinement end-to-end with the detection objective. CalibFusion builds a multi-frame persistence-aware Radar density representation with intensity weighting and Doppler-guided suppression of fast-varying clutter. A cross-modal transformer interaction module predicts a confidence-gated refinement of the initial extrinsics, which is integrated through a differentiable projection-and-splatting operator to generate calibration-conditioned image-plane Radar features.

Experiments on WaterScenes and FLOW show improved fusion-based 2D detection and robustness under synthetic miscalibration, supported by sensitivity analyses and qualitative Radar-to-image overlays. Results on nuScenes indicate that the refinement mechanism transfers beyond water-surface scenarios.

If accepted, we will release the implementation code.
\end{abstract}


%
\IEEEpeerreviewmaketitle

\section{Introduction}
Millimeter-wave (mmWave) Radar--Camera sensing is an attractive configuration for perception in autonomous and assisted navigation~\cite{10225711,9413181,Lin_2024_CVPR}. Cameras provide dense appearance cues for recognition and image-plane localization, while mmWave Radars offer complementary range--azimuth measurements that remain informative under low illumination and certain adverse weather. In Radar--Camera fusion, cross-modal aggregation depends on geometric consistency governed by extrinsic calibration~\cite{8917135,Luu_2025_CVPR}. When extrinsics deviate from their true values, Radar evidence is projected to biased image locations, cross-modal interactions become less reliable, and downstream tasks such as 2D detection can degrade~\cite{kim2023craft}.

Radar--Camera extrinsics are typically obtained offline using controlled setups or dedicated targets, but can drift over time due to mounting tolerances, vibration, thermal variation, or maintenance~\cite{11127186,10943662}. This motivates detectors that remain robust to miscalibration and adapt fusion when alignment cues are available.

Most existing calibration and auto-calibration methods are developed for road and urban environments~\cite{10225711,9413181}, where structured layouts and frequent objects provide correspondence constraints. Many approaches estimate extrinsics explicitly, e.g., by matching projected Radar observations with image features or associating image detections with Radar clusters~\cite{Luu_2025_CVPR,10943662}. In water-surface environments~\cite{yao2024waterscenes,guan2023achelous}, these cues are weaker: images may include large textureless regions, Radar returns are often sparse and intermittent with wave-/specular-induced clutter, and limited elevation information increases sensitivity to height assumptions. As a result, object-centric matching can be weakly constrained and per-frame explicit estimates may be unstable, undermining fusion-based detection.

We therefore embed Radar--Camera alignment into the fusion detector as a latent variable optimized for detection. Starting from $T_0$, CalibFusion predicts a sample-adaptive refinement and integrates it via a differentiable projection-and-splatting operator to generate calibration-conditioned image-plane Radar features, providing a gradient path from detection supervision to alignment variables. To improve cue reliability in sparse and cluttered water scenes, Radar observations are encoded as a multi-frame persistence-aware density with intensity weighting and Doppler-guided suppression of fast-varying returns. Cross-modal transformer interaction extracts soft correspondence patterns, and confidence-gated updates with temporal regularization improve stability when cues are insuffici

\section{Related Work}
\label{sec:related}
\subsection{Radar--Camera Calibration under Extrinsic Drift} Target-based offline calibration (e.g., checkerboards, corner reflectors, and carefully designed radar targets) can yield accurate radar--camera extrinsics under controlled conditions. However, in real deployments the effective extrinsics may deviate from the offline solution due to mounting tolerances, vibration, thermal variation, and maintenance operations, which motivates calibration and refinement mechanisms that can operate during regular sensing. Recent learning-based approaches typically estimate a 6-DoF correction by extracting features from radar and images and regressing an extrinsic update from cross-modal correlations~\cite{10943662,9981418}. Nevertheless, most existing benchmarks and designs focus on road and urban environments, where structured backgrounds and frequent objects provide persistent geometric cues~\cite{9561938}. In water-surface scenes, large texture-poor regions and intermittent, cluttered radar returns substantially weaken these constraints, making direct regression or correlation-based refinement less stable and less task-relevant.

\subsection{Object-Centric Association for Calibration} A widely adopted pragmatic paradigm is object-centric association: applying 2D detection on images, clustering radar measurements, and aligning the two modalities by matching object instances with heuristic criteria (e.g., box-center proximity, overlap after coarse projection, or temporal co-occurrence)~\cite{wang2023exploring}. This strategy can be effective when targets are dense and radar clustering is reliable. However, the association becomes poorly constrained when targets are sparse or radar returns are unstable, and the resulting calibration updates can be dominated by mismatches or transient clutter. These failure modes are particularly common in water-surface environments\cite{yao2024waterscenes,guan2023achelous,cheng2021flow}, where the number of salient objects per frame may be limited and radar observations may exhibit intermittency, making hard instance pairing brittle for continuous refinement.

\subsection{Calibration-Conditioned Fusion and Implicit Alignment in Detection} Beyond treating calibration as a standalone objective, a growing line of work integrates alignment into the downstream perception model, allowing cross-modal fusion to adapt to residual miscalibration. Attention-based fusion, especially transformer-style cross-modal interaction, is well suited to this setting because it supports soft, many-to-many correspondence learning rather than relying on hard matching~\cite{chen2022autoalign,WAN2026131914}. CalibFusion follows this implicit-alignment perspective by learning a bounded, confidence-gated extrinsic refinement that is optimized through the 2D detection objective. To improve the observability of alignment under sparse and cluttered radar observations, the method constructs a Doppler-guided multi-frame persistence density representation, and injects the refined extrinsics into fusion via a differentiable calibration-conditioned projection and splatting operator. This design ties extrinsic refinement directly to fusion detection performance and improves robustness when explicit association cues are weak.
\section{Methodology}
\label{sec:method}

\begin{figure*}[ht]
        \centering
        \includegraphics[width=0.9\linewidth]{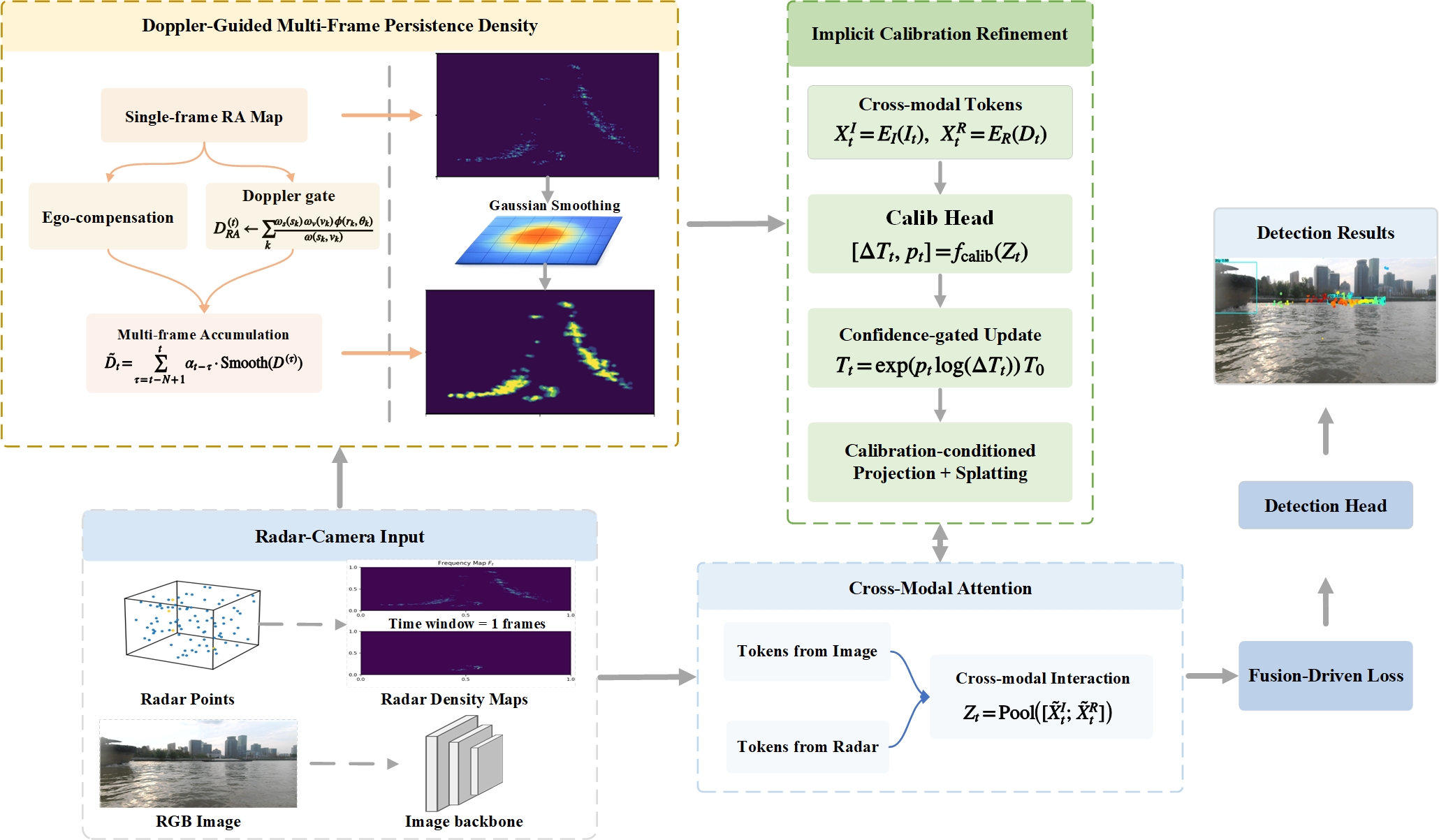}
\caption{\textbf{Overall architecture of CalibFusion.} CalibFusion includes: (1) Doppler-Guided Persistence Density for multi-frame Radar density construction; (2) Cross-Modal Token Interaction between image and Radar tokens; (3) Confidence-Gated Extrinsic Refinement that updates the initial extrinsic $T_0$; and (4) Calibration-Conditioned Projection that generates image-plane Radar features for fusion-based 2D detection. The model is trained end-to-end with detection supervision backpropagated through the differentiable projection.}
    \label{fig:main}
\end{figure*}

\subsection{Overview: Calibration-Conditioned Fusion Detection with Implicit Refinement}
\label{subsec:method_overview}
Fusion detection requires geometric consistency between Radar and camera. In water-surface deployments, Radar--camera extrinsics can drift from offline calibration due to vibration, thermal variation, or maintenance; small angular errors may cause pixel misprojection (approximately $f\cdot \Delta\theta$), affecting long-range and small targets. Radar returns are also sparse and subject to wave-induced clutter and intermittency, which weakens explicit object-level association.

As shown in Fig.~\ref{fig:main}, CalibFusion models extrinsic refinement as a latent alignment variable optimized by the detection objective. Starting from $T_0$, the refinement module predicts a sample-adaptive correction and conditions a differentiable projection to generate image-plane Radar features for fusion. Robustness is evaluated in Sec.~\ref{sec:experiments} using synthetic perturbations of $T_0$.

\paragraph{Confidence-gated extrinsic refinement.}
Let $T_0\in SE(3)$ denote the initial Radar-to-camera extrinsic transform. At time $t$, the refinement module predicts a corrective transform $\Delta T_t$ and a confidence score $\rho_t\in(0,1)$ indicating the reliability of the current alignment cues. The refined transform is obtained via a gated update in the Lie algebra:
\begin{equation}
\xi_t = \log(\Delta T_t)\in \mathbb{R}^6,\qquad
T_t = \exp(\rho_t\,\xi_t)\,T_0.
\label{eq:gated_extrinsic}
\end{equation}
The refined $T_t$ conditions the subsequent Calibration-Conditioned Projection, enabling end-to-end optimization where gradients from the detection loss propagate through the differentiable projection back to the refinement module.

\subsection{Doppler-Guided Persistence Density}
\label{subsec:seacalib_density}

Doppler-Guided Persistence Density constructs a temporally stable Radar density map $D_t$ under sparse and cluttered water-surface returns. $D_t$ is used to derive Radar tokens for Cross-Modal Token Interaction.

\paragraph{Single-frame range--azimuth density.}
A Radar detection at time $t$ is $p^{(t)}_{k}=(r^{(t)}_{k},\theta^{(t)}_{k},v^{(t)}_{k},s^{(t)}_{k})$. Detections are accumulated on a discretized range--azimuth grid via soft splatting:
\begin{equation}
D_{\mathrm{RA}}^{(t)}(i,j)\leftarrow D_{\mathrm{RA}}^{(t)}(i,j)+\omega\!\left(s^{(t)}_{k},v^{(t)}_{k}\right)\cdot \phi(r^{(t)}_{k},\theta^{(t)}_{k}; i,j),
\label{eq:ra_density_v2}
\end{equation}
where $\phi(\cdot)$ is a local interpolation kernel.

\paragraph{Doppler-guided weighting.}
The weight $\omega(\cdot)$ combines echo intensity and Doppler:
\begin{equation}
\omega(s,v)=w_s(s)\cdot w_v(v),
\label{eq:omega_factor}
\end{equation}
with
\begin{equation}
w_s(s)=\frac{\log(1+s)-\mu_s}{\sigma_s},\qquad w_s(s)\leftarrow \mathrm{clip}(w_s(s),0,1),
\label{eq:intensity_weight}
\end{equation}
and
\begin{equation}
w_v(v)=\exp\!\left(-\frac{v^2}{2\sigma_v^2}\right),
\label{eq:doppler_gate}
\end{equation}
where $\sigma_v$ controls the decay with Doppler magnitude. This weighting emphasizes returns that are more temporally stable for alignment while retaining the full Radar evidence for downstream feature extraction and fusion.

\paragraph{Multi-frame accumulation with persistence weighting.}
To mitigate intermittency, density maps are aggregated over a temporal window of $N$ frames:
\begin{equation}
\widetilde{D}_t = \sum_{\tau=t-N+1}^{t}\alpha_{t-\tau}\cdot \mathrm{Smooth}\!\left(D^{(\tau)}\right),
\label{eq:temporal_agg_v2}
\end{equation}
where $\alpha_{k}$ is a temporal weighting (e.g., exponential decay)
\begin{equation}
\alpha_k=\frac{\gamma^{k}}{\sum_{j=0}^{N-1}\gamma^j},\qquad k\in\{0,\ldots,N-1\},
\label{eq:alpha_decay}
\end{equation}
and $\mathrm{Smooth}(\cdot)$ is implemented as a depthwise Gaussian convolution:
\begin{equation}
\begin{aligned}
\mathrm{Smooth}(D) &= G_{\sigma_g} * D,\\
G_{\sigma_g}(\Delta i,\Delta j) &\propto
\exp\!\left(-\frac{\Delta i^2+\Delta j^2}{2\sigma_g^2}\right).
\end{aligned}
\label{eq:gaussian_smooth}
\end{equation}

A frequency map encodes persistence:
\begin{equation}
F_t(i,j)=\sum_{\tau=t-N+1}^{t}\mathbb{I}\!\left[D^{(\tau)}(i,j)>\epsilon\right],
\label{eq:freq_map_v2}
\end{equation}
and the final density combines aggregated intensity and persistence:
\begin{equation}
D_t(i,j)=\mathrm{Norm}\!\left(\widetilde{D}_t(i,j)\cdot g(F_t(i,j))\right),
\label{eq:final_density_v2}
\end{equation}
where
\begin{equation}
g(F)=\log(1+F)\quad \text{or}\quad g(F)=\left(\frac{F}{N}\right)^{\kappa},
\label{eq:persistence_weight}
\end{equation}
and $\mathrm{Norm}(\cdot)$ denotes per-map normalization:
\begin{equation}
\mathrm{Norm}(X)=\frac{X-\min(X)}{\max(X)-\min(X)+\varepsilon}.
\label{eq:normalize}
\end{equation}

\paragraph{Ego-motion compensation.}
If ego-motion $T_{\tau\rightarrow t}^{ego}$ is available, each past Radar detection is transformed into the current reference frame prior to accumulation:
\begin{equation}
\mathbf{X}^{\mathcal{R}}_{k,\tau\rightarrow t}=T_{\tau\rightarrow t}^{ego}\,\mathbf{X}^{\mathcal{R}}_{k,\tau}.
\label{eq:ego_comp}
\end{equation}
After transformation, $(r,\theta,v,s)$ are recomputed and $D^{(\tau)}$ is constructed accordingly, which improves spatial consistency across the accumulation window.

\subsection{Cross-Modal Token Interaction and Refinement Prediction}
\label{subsec:seacalib_xattn}

\begin{figure}[!t]
    \centering
    \includegraphics[width=\linewidth]{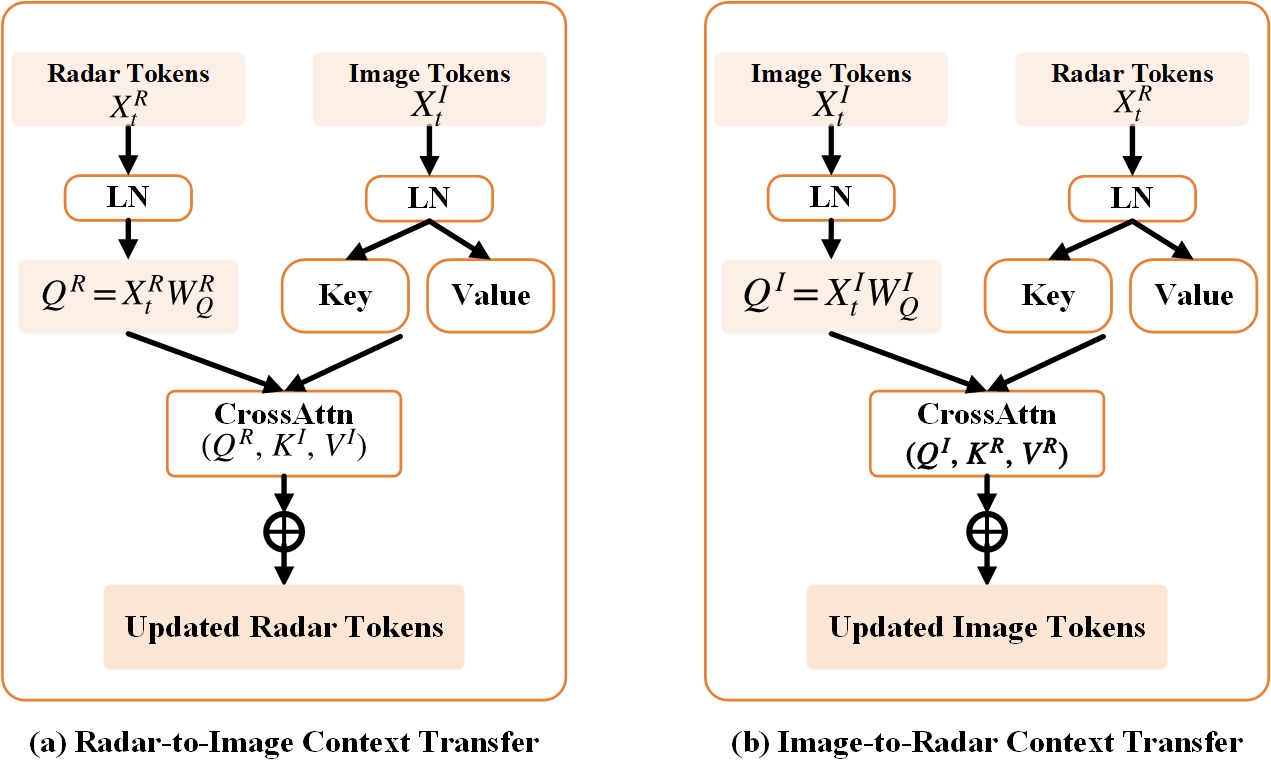}
\caption{\textbf{Cross-Modal Token Interaction via bi-directional cross-attention.}
(a) Radar-to-image attention updates Radar tokens using Radar queries and image keys/values. (b) Image-to-Radar attention updates image tokens using image queries and Radar keys/values. Layer normalization and residual updates are applied around each attention block.}
\label{fig:dual-stream}
\end{figure}

This section introduces Cross-Modal Token Interaction, which models Radar--image correspondence through token-level information exchange and provides the representation used by the refinement head, as illustrated in Fig.~\ref{fig:dual-stream}.

\paragraph{Tokenized encoding.}
Given the RGB image $I_t$, an image encoder $E_I$ produces visual tokens
\begin{equation}
\mathbf{X}^{I}_t = E_I(I_t)\in \mathbb{R}^{M\times d},
\label{eq:image_tokens_v2}
\end{equation}
and given the Radar density map $D_t$, a Radar encoder $E_R$ yields Radar tokens
\begin{equation}
\mathbf{X}^{R}_t = E_R(D_t)\in \mathbb{R}^{K\times d}.
\label{eq:Radar_tokens_v2}
\end{equation}
2D positional encodings $\mathbf{P}^I,\mathbf{P}^R$ are added:
\begin{equation}
\widetilde{\mathbf{X}}^I_t=\mathbf{X}^I_t+\mathbf{P}^I,\qquad
\widetilde{\mathbf{X}}^R_t=\mathbf{X}^R_t+\mathbf{P}^R.
\label{eq:pos_enc_v2}
\end{equation}

\paragraph{Bi-directional cross-attention.}
$L$ layers of bi-directional cross-attention are applied to exchange information between modalities. At layer $\ell$, Radar-to-image attention updates Radar tokens:
\begin{equation}
\mathbf{Z}^{R,\ell}_t=
\mathrm{softmax}\!\left(\frac{\mathbf{Q}^{R,\ell}_t(\mathbf{K}^{I,\ell}_t)^\top}{\sqrt{d}}+\mathbf{B}^{R\rightarrow I}\right)\mathbf{V}^{I,\ell}_t,
\label{eq:r2i_attn_v2}
\end{equation}
and image-to-Radar attention updates image tokens:
\begin{equation}
\mathbf{Z}^{I,\ell}_t=
\mathrm{softmax}\!\left(\frac{\mathbf{Q}^{I,\ell}_t(\mathbf{K}^{R,\ell}_t)^\top}{\sqrt{d}}+\mathbf{B}^{I\rightarrow R}\right)\mathbf{V}^{R,\ell}_t.
\label{eq:i2r_attn_v2}
\end{equation}
Residual updates with pre-norm are applied:
\begin{equation}
\begin{aligned}
\widetilde{\mathbf{X}}^{R,\ell+1}_t &=
\widetilde{\mathbf{X}}^{R,\ell}_t
+\mathrm{MLP}\!\left(\mathrm{LN}\!\left(\mathbf{Z}^{R,\ell}_t\right)\right),\\
\widetilde{\mathbf{X}}^{I,\ell+1}_t &=
\widetilde{\mathbf{X}}^{I,\ell}_t
+\mathrm{MLP}\!\left(\mathrm{LN}\!\left(\mathbf{Z}^{I,\ell}_t\right)\right).
\end{aligned}
\label{eq:residual_update_v2}
\end{equation}

After $L$ layers, the fused representation is pooled:
\begin{equation}
\mathbf{z}_t=\mathrm{Pool}\!\left([\widetilde{\mathbf{X}}^{R,L}_t;\widetilde{\mathbf{X}}^{I,L}_t]\right),
\label{eq:pool_v2}
\end{equation}
and used to predict the refinement parameters.

\paragraph{Refinement head.}
A lightweight head predicts rotation, translation, and confidence:
\begin{equation}
[\widehat{\mathbf{q}}_t,\widehat{\mathbf{t}}_t,\rho_t]=f_{\mathrm{calib}}(\mathbf{z}_t),\qquad
\widehat{\mathbf{q}}_t\leftarrow \frac{\widehat{\mathbf{q}}_t}{\|\widehat{\mathbf{q}}_t\|_2+\varepsilon}.
\label{eq:head_v2}
\end{equation}
The corrective transform is formed as $\Delta T_t=\mathrm{SE3}(\widehat{\mathbf{q}}_t,\widehat{\mathbf{t}}_t)$ and integrated via Eq.~\eqref{eq:gated_extrinsic}.

\subsection{Calibration-Conditioned Projection and Differentiable Splatting}
\label{subsec:seacalib_warp}

Calibration-Conditioned Projection converts Radar detections into an image-plane feature map conditioned on the refined extrinsic $T_t$, providing a differentiable path from the detection loss to the refinement variables.

\paragraph{From Radar detection to 3D point hypothesis.}
A 3D point is formed in the Radar frame on a reference plane with fixed height $h_0$:
\begin{equation}
\mathbf{X}^{\mathcal{R}}_k=
\begin{bmatrix}
r_k\cos\theta_k\\
r_k\sin\theta_k\\
h_0\\
1
\end{bmatrix}.
\label{eq:Radar_point_v2}
\end{equation}

\paragraph{Camera projection under refined extrinsic.}
Let $T_t=\begin{bmatrix}\mathbf{R}_t&\mathbf{t}_t\\ \mathbf{0}^\top&1\end{bmatrix}$ and camera intrinsics
$\mathbf{K}=\begin{bmatrix}f_x&0&c_x\\0&f_y&c_y\\0&0&1\end{bmatrix}$.
The Radar point is transformed to the camera frame:
\begin{equation}
\mathbf{X}^{\mathcal{C}}_k=
\mathbf{R}_t
\begin{bmatrix}
r_k\cos\theta_k\\
r_k\sin\theta_k\\
h_0
\end{bmatrix}
+\mathbf{t}_t,
\label{eq:to_camera_v2}
\end{equation}
and projected to pixel coordinates:
\begin{equation}
\tilde{\mathbf{u}}_k=
\mathbf{K}
\begin{bmatrix}
X^{\mathcal{C}}_k/Z^{\mathcal{C}}_k\\
Y^{\mathcal{C}}_k/Z^{\mathcal{C}}_k\\
1
\end{bmatrix},\qquad
\mathbf{u}_k=(u_k,v_k)=(\tilde{u}_k,\tilde{v}_k).
\label{eq:project_v2}
\end{equation}

\paragraph{Differentiable splatting with mass normalization.}
Each detection is mapped to a $C$-channel feature vector:
\begin{equation}
\mathbf{f}_k=\psi(p_k)\in\mathbb{R}^{C},
\label{eq:feat_embed_v2}
\end{equation}
and splatted to obtain $R_t$:
\begin{equation}
R_t(\mathbf{u})=\sum_{k} \kappa(\mathbf{u}-\mathbf{u}_k)\,\mathbf{f}_k,
\label{eq:splat_v2}
\end{equation}
where $\kappa(\cdot)$ is a differentiable kernel (bilinear splatting):
\begin{equation}
\kappa(\mathbf{u}-\mathbf{u}_k)=
\max(0,1-|u-u_k|)\cdot \max(0,1-|v-v_k|).
\label{eq:bilinear_kernel_v2}
\end{equation}
Kernel-mass normalization is applied:
\begin{equation}
\Gamma_t(\mathbf{u})=\sum_k \kappa(\mathbf{u}-\mathbf{u}_k),\qquad
\bar{R}_t(\mathbf{u})=\frac{R_t(\mathbf{u})}{\Gamma_t(\mathbf{u})+\varepsilon}.
\label{eq:mass_norm_v2}
\end{equation}
The calibrated Radar map $\bar{R}_t$ is then fed into the fusion network for calibration-conditioned multimodal fusion.

\subsection{Detection-Driven Objectives for Implicit Refinement}
\label{subsec:seacalib_loss}

The network is trained end-to-end with a 2D detection loss (classification and box regression). Gradients propagate to the refinement module through Calibration-Conditioned Projection in Sec.~\ref{subsec:seacalib_warp}. Additional regularizers are introduced to stabilize refinement without requiring extrinsic ground truth.

\paragraph{Extrinsic supervision in controlled settings.}
When reference extrinsics $T^\ast$ are available (e.g., synthetic perturbation experiments), rotation and translation losses are applied:
\begin{equation}
\mathcal{L}_{R}=1-\left|\left\langle \widehat{\mathbf{q}}_t, \mathbf{q}^\ast\right\rangle\right|,\qquad
\mathcal{L}_{t}=\left\|\widehat{\mathbf{t}}_t-\mathbf{t}^\ast\right\|_{1},
\end{equation}
\begin{equation}
\mathcal{L}_{\mathrm{calib}}=\lambda_R\mathcal{L}_R+\lambda_t\mathcal{L}_t.
\end{equation}

\paragraph{Small-update prior and temporal smoothness.}
The Lie algebra update $\xi_t=\log(\Delta T_t)$ is regularized by:
\begin{equation}
\mathcal{L}_{\mathrm{prior}} = \|\rho_t\,\xi_t\|_2^2,\qquad
\mathcal{L}_{\mathrm{smooth}} = \|\rho_t\,\xi_t-\rho_{t-1}\,\xi_{t-1}\|_1.
\label{eq:prior_smooth_rewrite}
\end{equation}

\paragraph{Query-conditioned attention consistency.}
Let $\mathcal{M}_t$ denote matched query--ground-truth pairs at time $t$. For each matched query $m\in\mathcal{M}_t$, denote normalized attentions over image and Radar tokens as $\mathbf{a}^{img}_{m,t}$ and $\mathbf{a}^{rad}_{m,t}$. The consistency term is:
\begin{equation}
\mathcal{L}_{\mathrm{q\_attn}}=
\sum_{m\in\mathcal{M}_t}\left(
\left\|\mathbf{a}^{img}_{m,t}-\mathbf{a}^{img}_{m,t-1}\right\|_1
+
\left\|\mathbf{a}^{rad}_{m,t}-\mathbf{a}^{rad}_{m,t-1}\right\|_1
\right).
\label{eq:qattn}
\end{equation}

\paragraph{Overall loss for the refinement module.}
The auxiliary objective is
\begin{equation}
\mathcal{L}=
\alpha\,\mathcal{L}_{\mathrm{calib}}
+\beta\,\mathcal{L}_{\mathrm{prior}}
+\gamma\,\mathcal{L}_{\mathrm{smooth}}
+\eta\,\mathcal{L}_{\mathrm{q\_attn}},
\label{eq:seacalib_loss_rewrite}
\end{equation}
where $\alpha=0$ when extrinsic supervision is unavailable. In practice, $\mathcal{L}$ is combined with the detector loss to learn implicit refinement jointly with fusion-based detection.
\section{Experiments}
\label{sec:experiments}
This section evaluates CalibFusion on (i) fusion-based 2D detection on water-surface datasets and (ii) extrinsic refinement under synthetic miscalibration with known offsets. Experiments follow Fig.~\ref{fig:main}: persistence density, cross-modal interaction for refinement prediction, and calibration-conditioned projection for fusion detection.

\subsection{Datasets, Miscalibration Protocol, and Evaluation Metrics}
\label{subsec:datasets_metrics}

\paragraph{Datasets.}
We use WaterScenes~\cite{yao2024waterscenes} (primary water-surface benchmark), FLOW~\cite{cheng2021flow} (transfer across inland-water scenes; Table~\ref{tab:flow_main_results}), and nuScenes~\cite{Caesar2019nuScenesAM} (generality via standard auto-calibration comparisons; Table~\ref{tab:nuscenes_autocalib_compare}).

\paragraph{Synthetic miscalibration.}
Perturbations are injected into $T_0$:
\begin{equation}
T_0' = \exp(\delta \xi)\, T_0,\qquad \delta\xi\in\mathbb{R}^6,
\label{eq:miscalib_protocol}
\end{equation}
where $\delta\xi$ contains three rotation and three translation components. Two ranges are used (R1/R2); multiple directions are sampled and results are reported as mean$\pm$std.

\paragraph{Metrics.}
\textbf{Detection.} We report mAP$_{50}$ and mAP$_{50:95}$, and additionally AP$_{\text{long}}$/AP$_{\text{small}}$ for water datasets (see supplementary). Robustness is summarized under R1/R2.
\textbf{Calibration.} With known $T^\ast$, we report rotation (geodesic on $SO(3)$) and translation (Euclidean) errors; Fig.~\ref{fig:miscalib_4panels} reports errors versus injected offsets.

\subsection{Compared Methods and Implementation Details}
\label{subsec:exp_methods}
\paragraph{Detection baselines.}
We compare \textit{Camera-only}, \textit{Radar-only}, and \textit{Fixed-fusion} (Radar+camera with fixed $T_0$); CalibFusion enables refinement and calibration-conditioned projection.

\paragraph{Calibration baselines.}
Under synthetic supervision, an \textit{Explicit SE(3) regression} baseline regresses $\Delta T_t$ from $(I_t, D_t)$. On nuScenes, we include standard auto-calibration baselines (Table~\ref{tab:nuscenes_autocalib_compare}).

\paragraph{Implementation details.}
CalibFusion uses Swin Transformer~\cite{liu2021swin} for images and PointNet++~\cite{qi2017pointnet++} for Radar, fused by the head in Sec.~\ref{sec:method}. Unless ablated, capacity is kept comparable. Backbone variants are reported in Sec.~\ref{subsec:exp_ablation}.

\subsection{Main Results}
\label{subsec:exp_main}

\paragraph{Fusion detection on FLOW dataset.}
Table~\ref{tab:flow_main_results} reports results under different sensor settings.
With Radar+camera inputs, CalibFusion reaches 95.3 mAP$_{50}$ and 47.1 mAP$_{50:95}$, outperforming RCFNet (93.2 / 44.7) by 2.1 mAP$_{50}$ and 2.4 mAP$_{50:95}$. Fusion also exceeds the best camera-only (85.6) and Radar-only (74.4) mAP$_{50}$, suggesting improved cross-modal complementarity in inland-water scenes.

\begin{table}[t]
\centering
\caption{Performance comparison on the FLOW dataset under different sensor settings.}
\label{tab:flow_main_results}
\begin{tabular}{l l c c}
\hline
\textbf{Performance} & \textbf{Sensor} & \textbf{mAP$_{50}$} & \textbf{mAP$_{50:95}$} \\
\hline
\multicolumn{4}{l}{\textbf{Camera-only}}\\
YOLOv8-N~\cite{10533619}      & Camera & 75.2 & 29.9 \\
YOLOv8-S~\cite{10533619}      & Camera & 79.1 & 32.9 \\
YOLOv8-M~\cite{10533619}      & Camera & 79.5 & 33.1 \\
YOLOv8-L~\cite{10533619}      & Camera & 75.6 & 34.0 \\
YOLOv8-X~\cite{10533619}      & Camera & 79.7 & 34.3 \\
YOLOv10-L~\cite{wang2024yolov10}     & Camera & 80.7 & 44.1 \\
YOLOv11-L~\cite{jocher2024ultralyticsyolo11}     & Camera & 83.9 & 47.3 \\
RT-DETR-R18~\cite{Zhao_2024_CVPR}   & Camera & 81.7 & 45.2 \\
RT-DETR-R50~\cite{Zhao_2024_CVPR}   & Camera & 83.0 & 46.6 \\
RT-DETR-R501~\cite{Zhao_2024_CVPR}  & Camera & 84.4 & 48.2 \\
USVRT-DETR~\cite{ZHANG2025122926}    & Camera & 85.6 & 49.6 \\
\hline
\multicolumn{4}{l}{\textbf{Radar-only}}\\
YOLOv8-N~\cite{10533619}      & Radar  & 72.5 & 39.4 \\
YOLOv8-S~\cite{10533619}      & Radar  & 69.1 & 38.8 \\
YOLOv8-M~\cite{10533619}      & Radar  & 74.4 & 42.1 \\
YOLOv8-L~\cite{10533619}      & Radar  & 71.0 & 39.3 \\
YOLOv8-X~\cite{10533619}      & Radar  & 70.9 & 39.7 \\
YOLOv10-L~\cite{wang2024yolov10}     & Radar  & 70.6 & 39.5 \\
YOLOv11-L~\cite{jocher2024ultralyticsyolo11}     & Radar  & 73.5 & 41.2 \\
\hline
\multicolumn{4}{l}{\textbf{Radar + Camera}}\\
RCFNet~\cite{10446880}        & Radar + Camera & 93.2 & 44.7 \\
\hline
\textbf{CalibFusion (ours)} & \textbf{Radar + Camera} & \textbf{95.3} & \textbf{47.1} \\
\hline
\end{tabular}
\end{table}
\paragraph{Generality on nuScenes auto-calibration.}
Table~\ref{tab:nuscenes_autocalib_compare} compares CalibFusion with representative auto-calibration methods under two perturbation ranges.
On R1 ($\pm10^\circ$, $\pm0.25$ m), CalibFusion achieves the lowest mean rotation error (0.354$^\circ$), reducing the strongest baseline in this table (CalibDepth: 0.807$^\circ$) by 56.2\%. On R2 ($\pm20^\circ$, $\pm1.5$ m), CalibFusion again yields the lowest mean rotation error (0.852$^\circ$), improving over the best listed baseline rotation mean (CalibDepth: 1.686$^\circ$) by 49.5\%. These results indicate that the refinement mechanism learned for calibration-conditioned fusion remains effective in road scenes under larger pose offsets.
For translation, CalibFusion is competitive but not always the best in this comparison (e.g., R1 mean translation 18.928 cm). This is consistent with the design of CalibFusion: the refinement is optimized to improve downstream fusion alignment rather than to recover physically exact extrinsics in every degree of freedom.

\begin{table*}[t]
\centering
\caption{Comparison with LiDAR--Camera-based and Radar--Camera-based auto-calibration methods on the nuScenes dataset.}
\label{tab:nuscenes_autocalib_compare}
\setlength{\tabcolsep}{6pt}
\begin{tabular}{llcccccccc}
\toprule
\multirow{2}{*}{Range} & \multirow{2}{*}{Methods}
& \multicolumn{4}{c}{Rotation ($^\circ$)}
& \multicolumn{4}{c}{Translation (cm)} \\
\cmidrule(lr){3-6}\cmidrule(lr){7-10}
& & Mean & Roll & Pitch & Yaw & Mean & X & Y & Z \\
\midrule
\multirow{6}{*}{R1 ($\pm 10^\circ$, $\pm 0.25$ m)}
& LCCNet-1   & 1.603 & 0.123 & 3.130 & 1.556 & 16.531 & 22.992 & 17.648 & 8.954 \\
& NetCalib2  & 1.205 & 0.387 & 2.289 & 0.941 & 12.297 & 12.532 & 12.076 & 12.284 \\
& CalibDepth & 0.807 & 0.390 & 0.345 & 1.686 & 12.608 & 12.860 & 12.250 & 12.715 \\
\cmidrule(lr){2-10}
& Coarse     & 2.035 & 0.581 & 1.519 & 4.004 & -- & -- & -- & -- \\
& Fine       & 1.692 & 0.442 & 0.939 & 3.695 & -- & -- & -- & -- \\
& \textbf{CalibFusion (ours)} & \textbf{0.354} & \textbf{0.102} & \textbf{0.117} & \textbf{0.842}
               & 18.928 & 18.936 & 17.487 & 20.361 \\
\midrule
\multirow{7}{*}{R2 ($\pm 20^\circ$, $\pm 1.5$ m)}
& LCCNet-3   & 2.156 & 1.526 & 2.364 & 2.579 & 89.672 & 71.660 & 89.605 & 107.751 \\
& LCCNet-5   & 1.898 & 0.919 & 2.314 & 2.461 & 88.302 & 74.216 & 85.239 & 105.450 \\
& NetCalib2  & 2.778 & 1.465 & 4.688 & 2.180 & 71.037 & 76.001 & 57.204 & 79.906 \\
& CalibDepth & 1.686 & 1.149 & 0.808 & 3.102 & 55.380 & 77.146 & 12.918 & 76.078 \\
\cmidrule(lr){2-10}
& Coarse     & 4.388 & 1.866 & 3.251 & 8.048 & -- & -- & -- & -- \\
& Fine       & 3.334 & 1.368 & 1.937 & 6.696 & -- & -- & -- & -- \\
& \textbf{CalibFusion (ours)} & \textbf{0.852} & \textbf{0.302} & \textbf{0.395} & \textbf{1.239}
               & 75.459 & 75.156 & 82.631 & 85.1483 \\
\bottomrule
\end{tabular}
\end{table*}

\subsection{Qualitative Results}
\label{subsec:exp_qualitative}

\begin{figure}[!t]
    \centering
    \includegraphics[width=\linewidth]{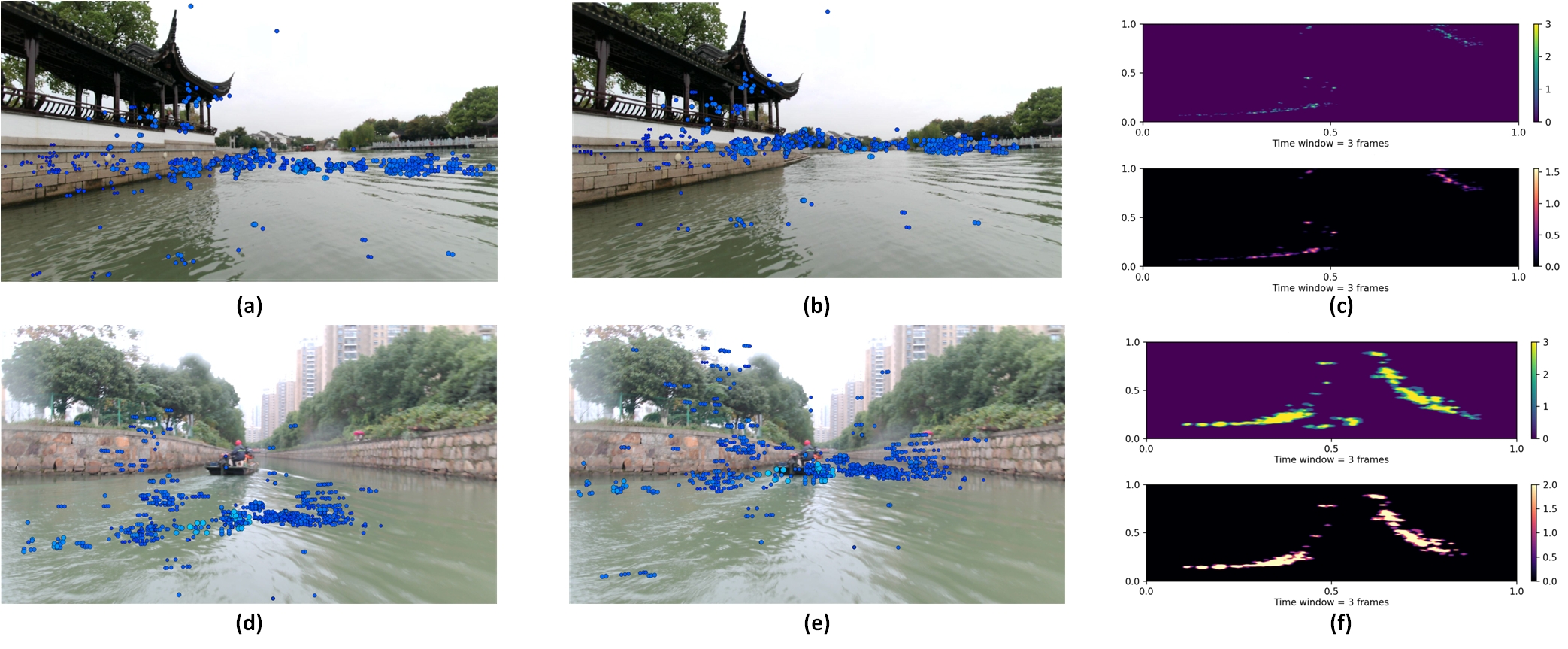}
    \caption{Calibration-conditioned projection results.
(a,d) Miscalibrated Radar-to-image projections under $T_0'$.
(b,e) Refined projections after CalibFusion predicts $T_t$, improving cross-modal alignment.
(c,f) Corresponding multi-frame Radar density maps used for refinement.}
    \label{fig:overlay}
\end{figure}

Fig.~\ref{fig:overlay} visualizes calibration-conditioned projection under synthetic miscalibration. 
Compared to the miscalibrated projection under $T_0'$, the refined transform $T_t$ yields Radar-to-image overlays that better align with image content. 
The density maps in (c,f) correspond to the Doppler-guided multi-frame persistence representation used by the refinement module.

\begin{figure}[!t]
    \centering
    \includegraphics[width=\linewidth]{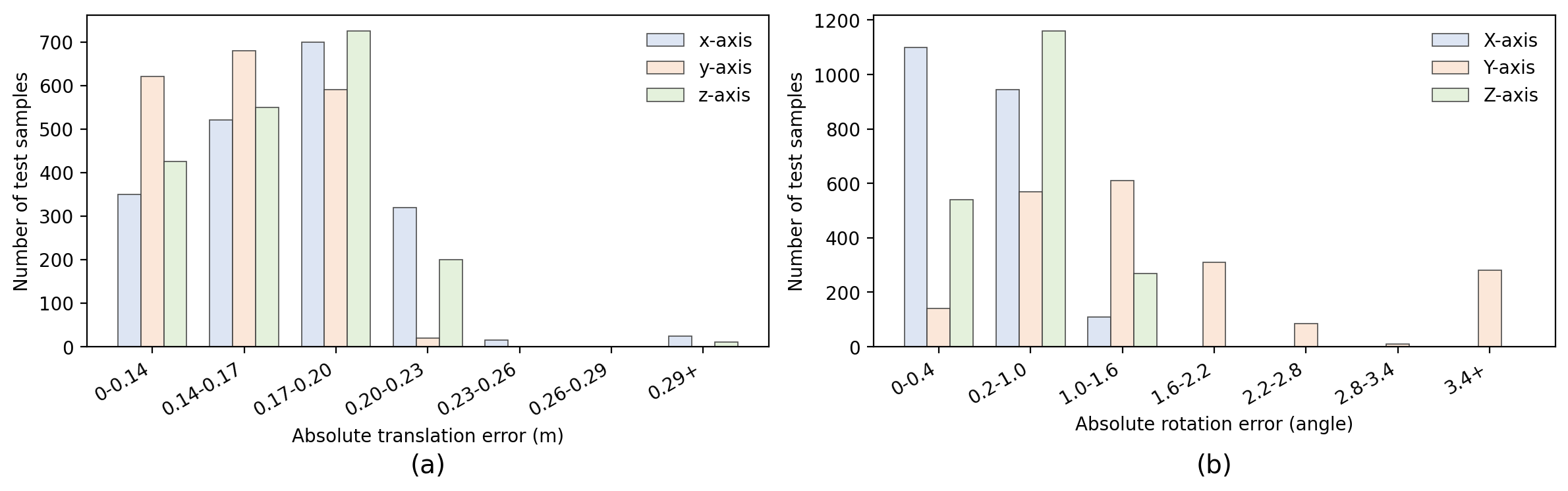}
    \caption{Calibration error statistics on the WaterScenes test set under a broad range of initial perturbations: (a) distribution of translation errors; (b) distribution of rotation errors.}
    \label{fig:dual_hist}
\end{figure}

\begin{figure}[!t]
    \centering
    \includegraphics[width=\linewidth]{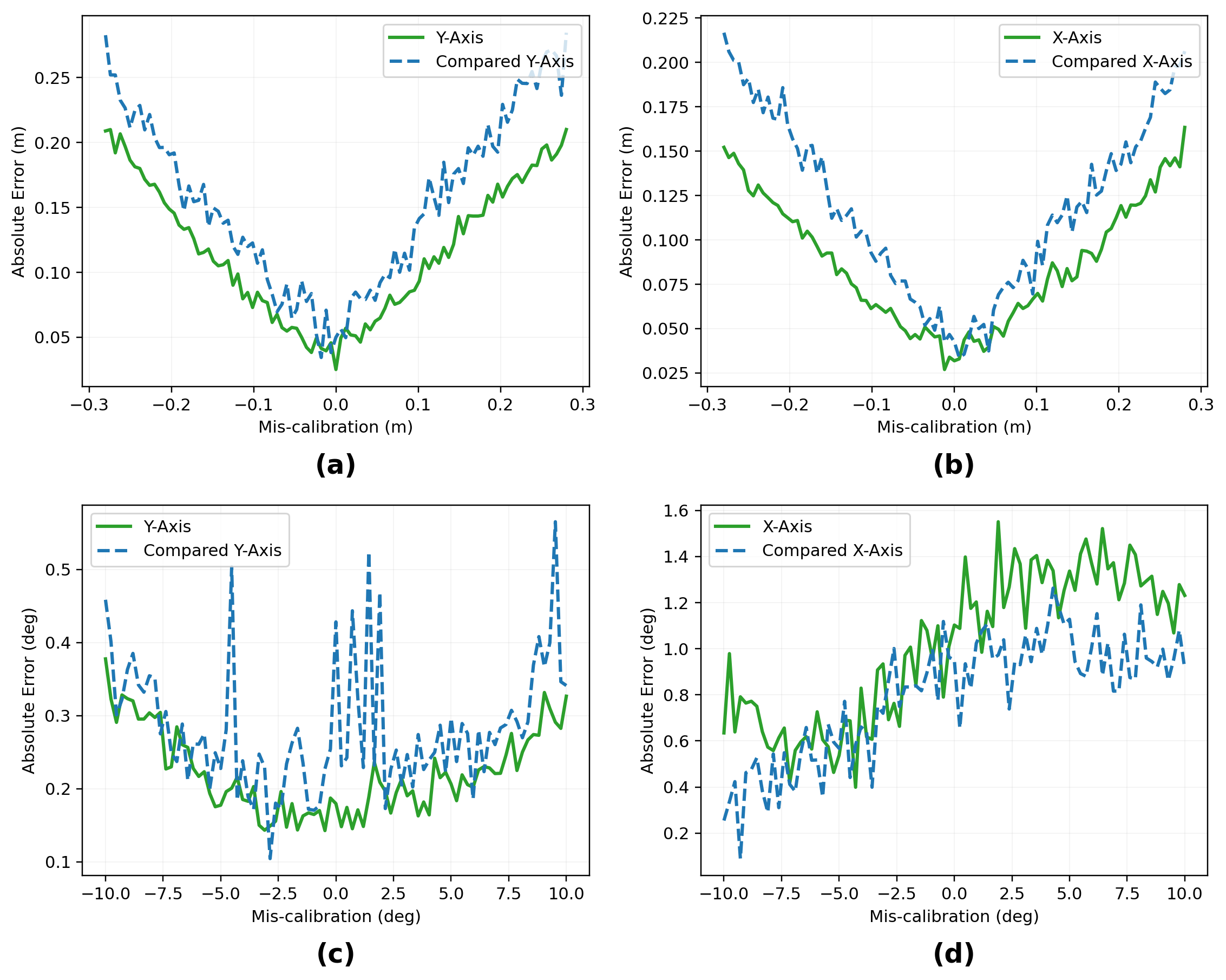}
    \caption{Refinement error under axis-wise injected extrinsic offsets. Top row: translation error (m); bottom row: rotation error (deg). In each subplot, the solid curve corresponds to the perturbed axis (X or Y), while the dashed curve (“Compared”) reports the error on the other axis to indicate cross-axis coupling. (a,c) Y-axis offsets; (b,d) X-axis offsets.}
    \label{fig:miscalib_4panels}
\end{figure}

To further examine the behavior of refinement under a broad range of injected offsets, 
Fig.~\ref{fig:dual_hist} summarizes the error distributions on the WaterScenes test set. 
Fig.~\ref{fig:miscalib_4panels} further reports absolute translation/rotation errors as the injected miscalibration increases along each axis. 
Across axes and magnitudes, the refined estimate reduces error relative to the perturbed initialization, which is consistent with the confidence-gated update.

\subsection{Qualitative Detection Results}
\label{subsec:exp_qualitative}

\begin{figure}[!t]
    \centering
    \includegraphics[width=\linewidth]{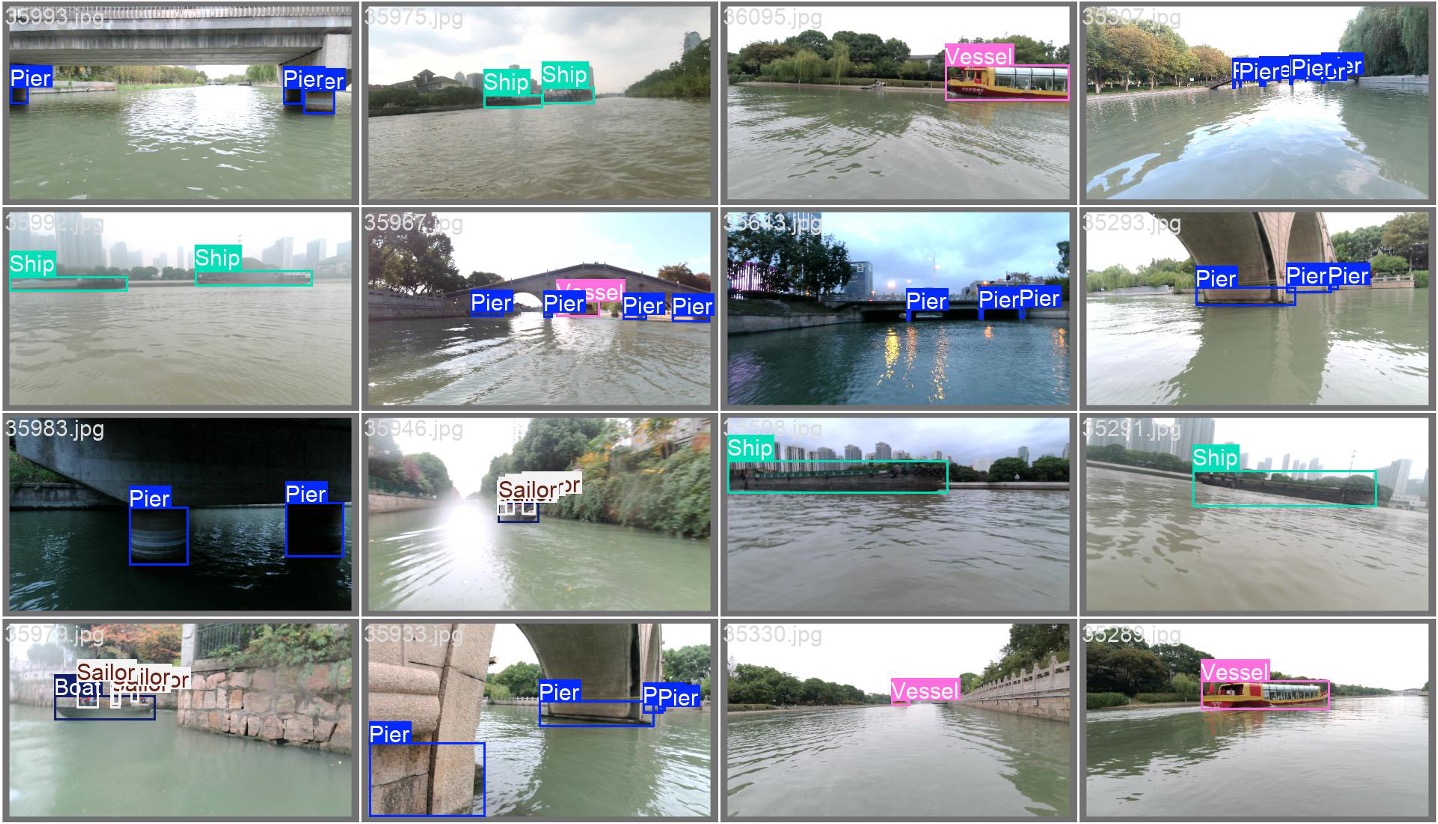}
    \caption{Representative detection results of CalibFusion on WaterScenes, illustrating Radar--Camera fusion detection across typical targets and water-surface environments.}
    \label{fig:qual_det}
\end{figure}

Fig.~\ref{fig:qual_det} provides representative qualitative detection results on WaterScenes. The examples illustrate typical targets and scenes where calibration-conditioned fusion integrates Radar evidence with visual cues for 2D detection.

\subsection{Ablation settings.}
\label{subsec:exp_ablation}
We perform ablations and compare image/Radar backbones. Table~\ref{tab-abl-FPN} shows that Swin Transformer (image) with PointNet++ (Radar) performs best, achieving 95.3 mAP$_{50}$ and 47.1 mAP$_{50:95}$. Replacing PointPillar with PointNet++ yields consistent mAP$_{50}$ gains (ResNet101: 94.2$\rightarrow$94.5~\cite{Szegedy_Ioffe_Vanhoucke_Alemi_2017}; SwinT.: 94.7$\rightarrow$95.3) with minor changes in mAP$_{50:95}$ (46.6--47.1). We therefore use Swin Transformer and PointNet++ as the default setting.
      \begin{table}[h]
		\centering
		\caption{Ablation study of image backbone and Radar backbone (Image Bb. )}
		\label{tab-abl-FPN}
		\small
			\begin{tabular}{c c c c}
\hline
 Image Backbone.&  Radar Backbone& mAP$_{50}$ & mAP$_{50:95}$\\ \hline
ResNet101& PointPillar & 94.2&  46.8 \\ \hline
ResNet101& PointNet++ & 94.5 &  46.7 \\ \hline
SwinT.& PointPillar & 94.7 &  46.6 \\ \hline
SwinT.& PointNet++ & \textbf{95.3} &  \textbf{47.1} \\ \hline
\end{tabular}
	\end{table}

\section{Conclusion}
\label{sec:conclusion}

We presented \textit{CalibFusion}, a calibration-conditioned Radar--Camera fusion framework that integrates implicit extrinsic refinement into a 2D detection pipeline. CalibFusion treats alignment as a latent variable optimized by the detection objective. It builds a multi-frame persistence-aware Radar density representation with intensity weighting and Doppler-guided suppression of fast-varying clutter, predicts a confidence-gated SE(3) refinement via cross-modal transformer interaction, and injects the refined transform through a differentiable projection-and-splatting operator for end-to-end learning.

Experiments on WaterScenes and FLOW show improved fusion-based 2D detection and increased robustness under synthetic miscalibration across rotation and translation offsets, supported by sensitivity curves and projection overlays. Results on nuScenes suggest that the refinement mechanism transfers beyond water-surface environments.

Future work will address height ambiguity and platform dynamics, improve temporal modeling for long-horizon drift, and validate on additional sensor configurations and real-world perturbations.


\section*{Acknowledgment}

Acknowledgements are omitted for double-blind review.

\bibliographystyle{IEEEtran}
\bibliography{ref}

\end{document}